\documentclass[a4paper,twoside]{article}

\usepackage{epsfig}
\usepackage{subcaption}
\usepackage{calc}
\usepackage{amssymb}
\usepackage{amstext}
\usepackage{amsmath}
\usepackage{amsthm}
\usepackage{multicol}
\usepackage{pslatex}

\usepackage{algorithm2e}
\usepackage{wrapfig}
\usepackage{graphicx}
\usepackage{booktabs}
\usepackage[bottom]{footmisc}
\usepackage[pagebackref,breaklinks,colorlinks]{hyperref}

\usepackage[capitalize]{cleveref}
\crefname{section}{Sec.}{Secs.}
\Crefname{section}{Section}{Sections}
\Crefname{table}{Table}{Tables}
\crefname{table}{Tab.}{Tabs.}
\usepackage[round]{natbib}
\usepackage{SCITEPRESS}     

\newcommand{\eg}{\textit{e.g.}\ }

\begin{document}

\title{Multi-task Planar Reconstruction with Feature Warping Guidance}

\author{\authorname{Luan Wei\sup{1},
Anna Hilsmann\sup{1} and Peter Eisert\sup{1,}\sup{2}}
\affiliation{\sup{1}Fraunhofer Heinrich Hertz Institute, Berlin, Germany}
\affiliation{\sup{2}Humboldt University, Berlin, Germany}
\email{\{luan.wei, anna.hilsmann, peter.eisert\}@hhi.fraunhofer.de}
}



\abstract{Piece-wise planar 3D reconstruction simultaneously segments plane instances and recovers their 3D plane parameters from an image, which is particularly useful for indoor or man-made environments. Efficient reconstruction of 3D planes coupled with semantic predictions offers advantages for a wide range of applications requiring scene understanding and concurrent spatial mapping. However, most existing planar reconstruction models either neglect semantic predictions or do not run efficiently enough for real-time applications. We introduce SOLOPlanes, a real-time planar reconstruction model based on a modified instance segmentation architecture which simultaneously predicts semantics for each plane instance, along with plane parameters and piece-wise plane instance masks. We achieve an improvement in instance mask segmentation by including multi-view guidance for plane predictions in the training process. This cross-task improvement, training for plane prediction but improving the mask segmentation, is due to the nature of feature sharing in multi-task learning. Our model simultaneously predicts semantics using single images at inference time, while achieving real-time predictions at 43 FPS. Code is available at: \url{https://github.com/fraunhoferhhi/SOLOPlanes}.}

\onecolumn \maketitle \normalsize \setcounter{footnote}{0} \vfill

\section{\uppercase{Introduction}}
\label{sec:intro}

Estimating the 3D structure of a scene holds importance across a variety of domains, including robotics, virtual reality (VR), and augmented reality (AR). The demand for real-time applications in these areas increases over time as such technologies proliferate. Man-made architectures and indoor environments, where the application end-users spend a significant amount of time, often consist of regular structures like planar surfaces, aligning well with the Manhattan world assumption that such surfaces typically exist on a regular 3D grid \citep{manhattan}. Estimating plane parameters directly can reduce noise for areas lying on a planar surface, which can be particularly useful for indoor scenes dominated by planar surfaces. It also holds relevance for outdoor scenarios, such as self-driving cars and outdoor AR applications, where streets and buildings often adhere to similar geometric principles.
\begin{figure}[!t]
    \centering
    \begin{subfigure}{0.494\linewidth}
    \includegraphics[width=\linewidth]{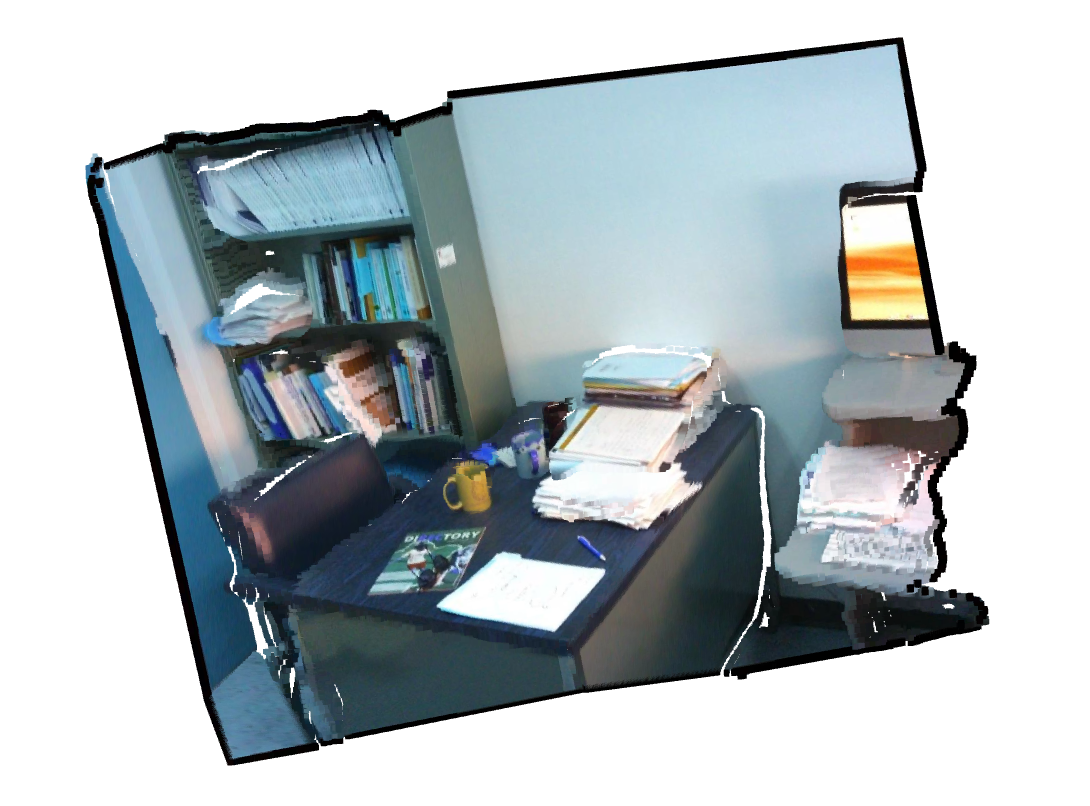}
          \end{subfigure}
              \begin{subfigure}{0.494\linewidth}
    \includegraphics[width=\linewidth]{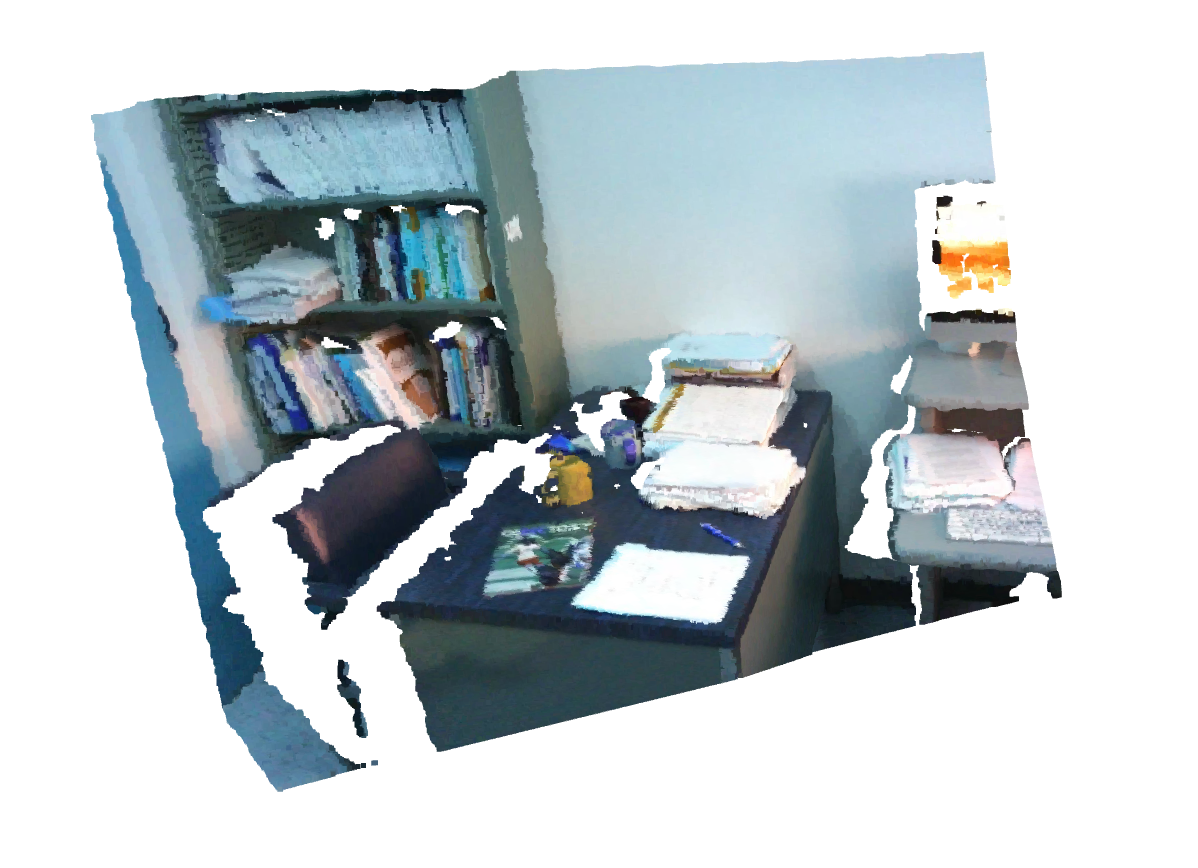}
          \end{subfigure}\\
    \begin{subfigure}{0.494\linewidth}
    \includegraphics[width=\linewidth]{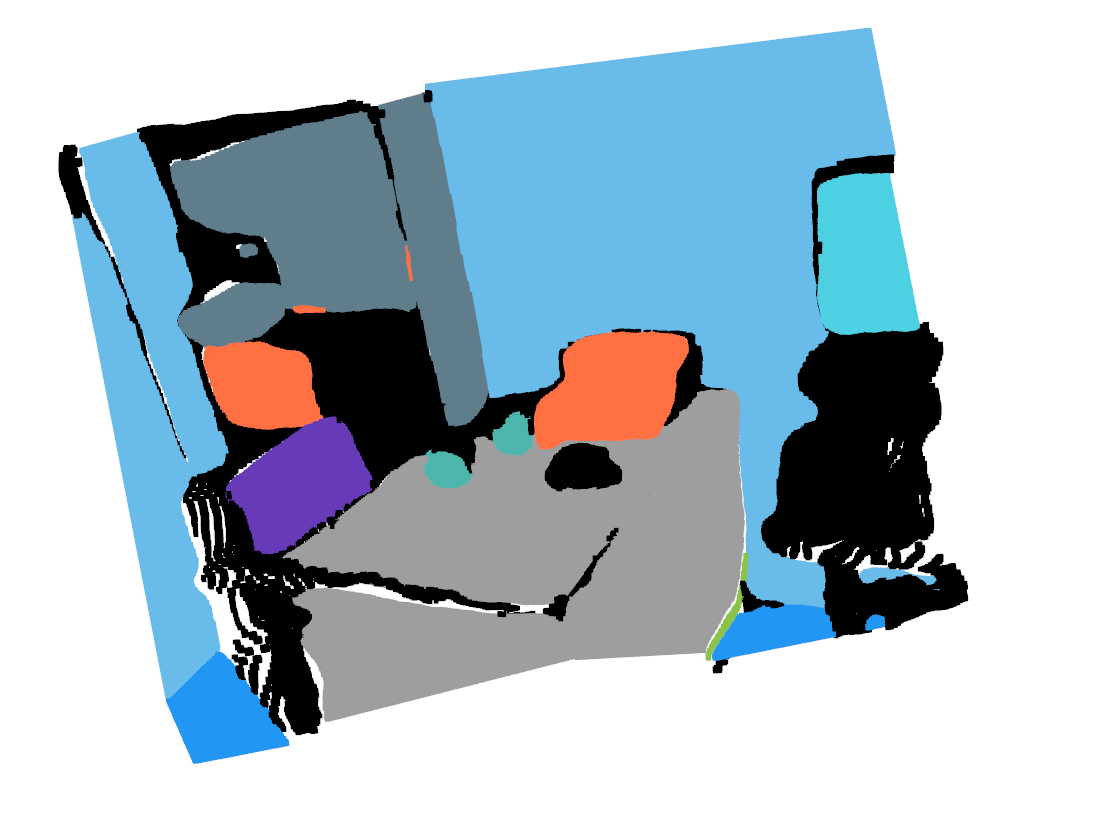}
    \caption{Ours}
          \end{subfigure}
    \begin{subfigure}{0.494\linewidth}
    \includegraphics[width=0.95\linewidth]{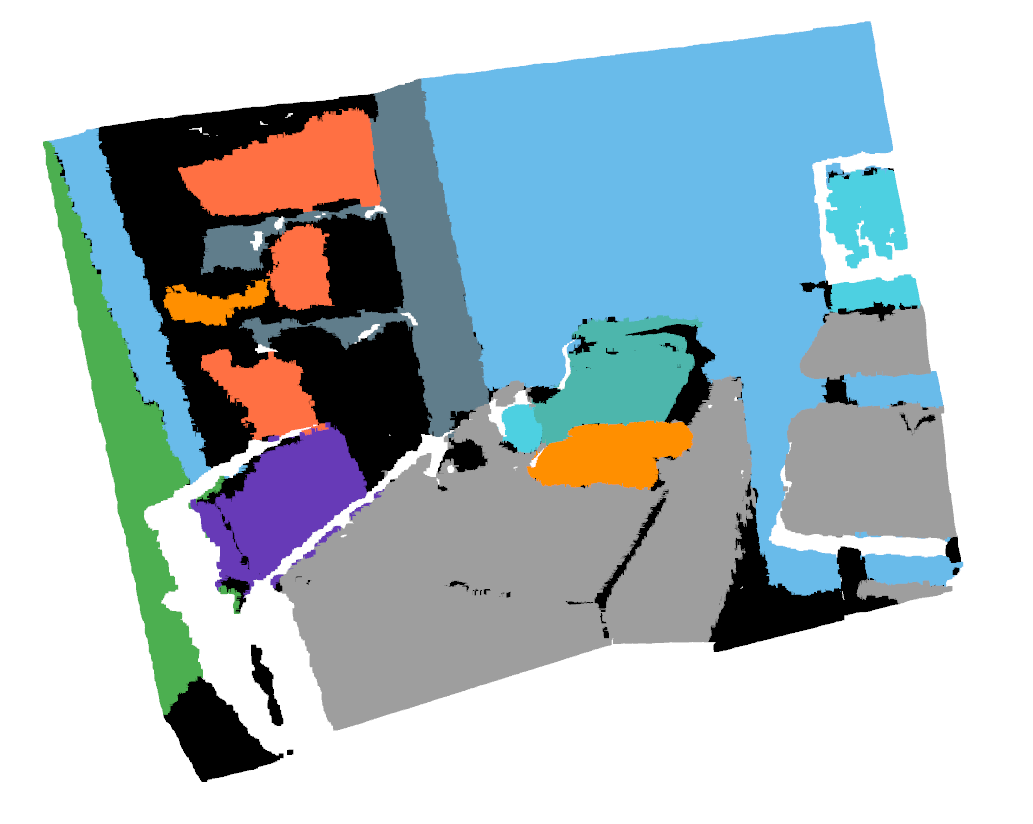}
    \caption{GT}
          \end{subfigure}
    \\[0.3em]
    \caption{Comparison of SOLOPlanes output with ground truth (GT). 3D projections using predicted plane parameters (left) and GT depth (right). Textures use RGB  (top), predicted semantics (bottom left), and GT semantics (bottom right).}
    \label{fig:teaser}
\end{figure}

Several methods have been proposed to use deep learning to recover planes of indoor scenes from a single image \citep{planenet, planeAE, planercnn, planerecnet, planesegnet}. While existing works have made strides in predicting piece-wise instance masks and plane parameters, they often ignore the added information from scene semantics. Incorporating semantics provides an added layer of scene understanding, which can be useful in many applications. For instance, the semantic label for a planar surface can help a service robot in determining the correct behaviour (\eg mopping floor vs. wiping table), or AR/VR experiences could offer semantics-dependent retexturization. Some models predict semantics along with plane parameters but are often too computationally intensive to meet the real-time requirements of practical applications \citep{planemvs}.

Multi-task learning, the technique of using a single model to learn multiple tasks concurrently, has shown promise in terms of data efficiency and improved generalization \citep{multitask1997}. However, recent studies indicate that there is also added difficulty in jointly learning multiple tasks. While some tasks may benefit from being learned together, thereby boosting accuracy, others may interfere with each other, leading to worse performance \citep{whichtasks-mtl2020}.

Our aim was to create a data-efficient model with improved run-time efficiency compared to existing models for planar reconstruction with semantics. We achieve the desired outcome via our model, SOLOPlanes (SOLOP), where we make use of multi-view guidance for improved data usage when acceptable ground truth plane segments differ across views, and made adjustments to the base architecture for improved efficiency. Multi-view warping is done in feature space, by warping plane features from neighbour to source view, decoding, then transforming the decoded plane parameters to the source view camera view for comparison with ground truth data during training. This additional warping guidance for plane features positively impacts the learning of segmentation masks, particularly when using a more limited dataset, while only requiring a single view at inference time.

In the context of our work, we found that multi-view guidance using plane features leads to a notable improvement in segmentation results. We attribute this enhancement to our multi-task architecture and the use of a shared trunk, meaning a global feature extractor that is common to all tasks \citep{mtl-survey20}. This architecture allows for loss propagation through shared features and common base networks, and may be particularly relevant in the case of incomplete or varying data across overlapping views.\\

\noindent Our contributions include the following: 
\begin{enumerate}
    \item An empirical demonstration of cross-task improvement using multi-view guidance by feature warping, with particular relevance in cases where ground truth data may be incomplete across neighboring views. 
    \item A single-image planar reconstruction model, that can concurrently predict semantics for planar segments while achieving the best efficiency compared to other known planar reconstruction methods at a processing speed of 43 FPS. 
\end{enumerate}
Our approach may be a helpful method for other multi-task models limited in some forms of ground truth training data. The efficiency of the model makes it suitable for a range of real-world applications.


\section{\uppercase{Related Work}}
\label{sec:related}

\textbf{Planar Reconstruction} Early works in planar reconstruction using a single image predicted a set number of planes per scene without using ground truth plane annotations by employing a plane structure-induced loss \citep{planes18}. Another early end-to-end planar reconstruction network from a single image is PlaneNet, which uses separate branches for plane parameter, mask, and non-planar area depth estimation \citep{planenet}. Two major subsequent models serve as the foundation for several later works. PlaneRCNN is an extension of the two-stage instance segmentation model, Mask-RCNN \citep{maskrcnn}, and predicts the plane instance normal and depth map, then jointly process plane parameters along with segmentation masks through a refinement module \citep{planercnn}. PlaneAE predicts per-pixel parameters and associative embeddings and employs efficient mean clustering to group the pixel embeddings to plane instances \citep{planeAE}. PlaneTR uses geometric guidance by generating and tokenizing line segments, giving the input additional structural information \citep{planetr}. Additional contributions in this area include post-processing refinement networks that enforce interplane relationships, via predicting the contact line or geometric relations between adjacent planes \citep{interplanes}. More recent works follow the approach of using an instance segmentation model base. PlaneSegNet is based on a real-time instance segmentation architecture and introduces an efficient Non-Maximum Suppression (NMS) technique to reduce redundant proposals \citep{planesegnet}. PlaneRecNet predicts per-pixel depth and plane segmentation masks, then use classical methods like PCA or RANSAC to recover plane parameters \citep{planerecnet}. A number of single-image plane reconstruction models use some form of instance segmentation baseline. However, most single-image models focus solely on spatial parameters and largely ignore the task of recovering semantics.

\textbf{Multi-view approaches} The task of predicting 3D plane parameters from a single image is inherently ambiguous and challenging. Thus, several works have incorporated multi-view information, either as a loss guidance or by using multiple image inputs at inference time. PlanarRecon \citep{planarrecon} is a real-time model using multiple image frames which makes predictions directly in 3D by using a coarse-to-fine approach for building a sparse feature volume, then clustering occupied voxels for instance planes, and uses a tracking and fusion module to get a global plane representation. PlaneMVS \citep{planemvs} is the first to apply a deep multi-view stereo (MVS) approach to plane parameters. Although it achieves state-of-the-art results and also predicts class semantics, it is less computationally efficient due to the use of 3D convolutions and requires generation of plane hypotheses. PlaneRCNN incorporates a multi-view warping loss module that enforces consistency with nearby views by projecting the predictions to 3D and calculating the distance after transforming to the same camera coordinates \citep{planercnn}. Unlike our approach, their warping module is applied directly on the predictions rather than in feature space. Another work enhances the PlaneAE model with multi-view regularization by warping the plane embedding feature maps and using associative embeddings from multiple views to improve the final instance segmentation \citep{mvreg-planeae-Xi_Chen_2019}. 


\textbf{Feature Warping} Feature warping is commonly done in deep Multi-View Stereo (MVS) approaches, as it was found that creating the cost volume using features is as effective for artificial neural networks and more computationally efficient due to reduced size \citep{dpsnet, yao2018mvsnet}. While some approaches use a similarity function on the features, others simply concatenate the warped feature with the original and let the model learn the relation rather than calculate an explicit cost volume \citep{chen2020visibility, yao2018mvsnet}. The latter approach is used by PlaneMVS to construct a Feature/Cost volume, which is then processed by a 3D CNN to get the plane parameters. Deep MVS methods are more commonly used for depth estimation, and their application to plane parameter estimation is relatively novel. Other research suggests that calculating a feature error between frames is more robust than a photometric error \citep{depthcamfeatwarp21}. However, this cannot directly be applied to plane reconstruction, as the plane features contain information in different camera views when considering a video dataset. Takanori et al.~use multi-frame attention via feature warping for the task of drone crowd tracking \citep{dronefeatwarp23}. Ding et al.~take MVS as a feature matching task and use a Transformer model to aggregate long-range global context using warped feature maps \citep{transmvsnet_2022_CVPR}. 

In order to ensure differentiability, the warp to another view using depth values and camera parameters must be backprojected using bilinear interpolation. Most existing works involving feature warping do not specifically deal with plane features, which require transformation to the correct view when decoded. Additionally, the majority of planar reconstruction models do not offer semantic predictions for the scene.

The majority of existing works primarily focus on the geometric accuracy of planes without holistically addressing the more practical requirements of speed and semantic understanding of planar scenes. Our work aims to fill this gap by offering a unified framework for semantic planar reconstruction. We improve data efficiency during training and achieve cross-task improvement using multi-view guidance for plane features, while maintaining an inference speed that is suitable for real-time applications.

\section{\uppercase{Method}}
\label{sec:Method}
Our objective is to develop a real-time framework for the task of 3D semantic planar reconstruction. This section is organized as follows: Section \ref{sec:overview} provides details of the framework, Section \ref{sec:losses} elaborates on the loss terms, and Section \ref{sec:mvfeatwarp} introduces our multi-view guidance. 

\subsection{Framework}
\label{sec:overview}

Our framework is built on a light version of the SOLOv2 instance segmentation model \citep{solov2} using a ResNet-50 backbone \citep{resnet}. SOLOv2 is a single-stage instance segmentation model that predicts instance masks and labels based on their spatial location. It achieves an execution speed of around 31 FPS using a single V100 GPU card \citep{solov2}. The model employs dynamic convolution to generate the final segmentation mask, leveraging multi-scale features from the Feature Pyramid Network (FPN) \citep{fpn}. Each level of the FPN output features are used to predict mask kernels and class semantics, with the features reshaped to square grids of varying sizes, with each responsible for predictions at a different scale. Each grid location predicts a kernel and semantic category scores. The mask feature is obtained through feature fusion of the first four levels of the FPN outputs via bilinear upsampling and convolution layers, and the final segmentation masks are obtained via convolution using the predicted kernels, with redundant masks suppressed using matrix Non-Maximum Supression (NMS) \citep{solov2}. The mask and kernel features receive spatial awareness information from concatenated normalized coordinate, a method from \cite{coordconv}.

\begin{figure}[!h]
\centering
\includegraphics[width=0.95\linewidth]{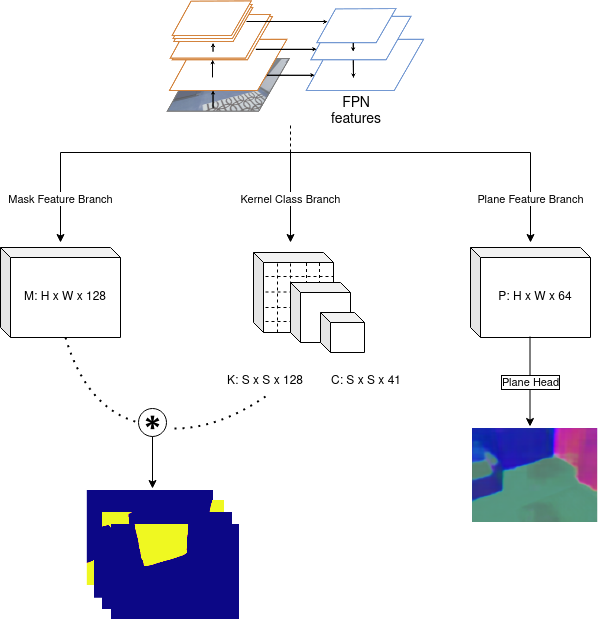} 
\caption{Simplified overview of SOLOPlanes architecture.}
\label{fig:arch}
\end{figure}

\vspace{1em}
We extend the base architecture by introducing a plane feature branch that fuses the first two levels of the feature map, along with a plane prediction head that outputs per-pixel plane parameters via a convolution layer (see Fig. \ref{fig:arch}). This prediction is supervised by a set of loss functions that leverage geometrical constraints and ground truth depth information (detailed in Section \ref{sec:losses}).
The original architecture predicts the kernels and semantic categories using all five feature map levels of the FPN. Based on the findings of \cite{yolof}, a divide-and-conquer approach is more crucial than leveraging multi-scale features for task-specific predictions, we experimented with using different feature levels and found that using fewer feature levels not only maintained comparable performance in multi-task planar segmentation but also improved the overall efficiency of the model.

Our final architecture takes a single RGB image, $I \in \mathbb{R}^{H\times W\times C}$, as input during inference, and outputs an arbitrary number of plane instance masks along with instance level semantics and per-pixel plane parameters. We obtain the final result by pooling per-pixel parameter prediction using the predicted masks, and retaining per-pixel predictions in areas without a plane instance. The model is trained on the large-scale public ScanNet dataset containing indoor scenes from \cite{scannet}, supplemented with ground truth plane annotations from \cite{planercnn}.

\subsection{Losses}
\label{sec:losses}
\textbf{Mask \& Category} We retain the original loss functions from \cite{solov2} for mask and category predictions. The Dice Loss, $L_M$, guides mask prediction with the original loss weight $w_M = 3$, and focal loss, $L_C$, for semantic category prediction \citep{focalloss}. For full details, we refer readers to \citep{solo}. 
In order to address class imbalances due to dominating negative samples, we modified $L_C$ to only consider grid locations containing an instance. 

\textbf{Plane Parameters} Plane parameters are represented by the normal and offset of the plane, denoted as $\mathbf{p} = (\mathbf{n}, d)$, which we combine into a single parameter $\mathbf{p} = \mathbf{n}*d \in \mathbb{R}^3$, with $\mathbf{n}$ normalized to unit length. Due to the complexity of predicting plane parameters containing both normal and depth information, we employ multiple loss functions for supervising per-pixel plane predictions. We use L$_1$ loss for direct comparison with ground truth plane parameters:  
\begin{equation}
    L_{plane} = \frac{1}{N} \sum_{i=1}^{N} \Vert \mathbf{p}_i - \mathbf{p}_i^* \Vert.
\end{equation}

An asterisk is used to denote predicted values, and $N$ represents the total number of pixels. The cosine distance, denoted $L_{surface}$, is used to guide the learning of surface normals. Due to the way we represent plane parameter $\mathbf{p}$, we get the equivalent result calculating cosine similarity on plane parameters directly. 
\begin{equation}
    sim_i = \frac{\mathbf{p}_i\cdot \mathbf{p}_i^* }{\Vert \mathbf{p}_i\Vert \Vert \mathbf{p}_i^*  \Vert}, \hspace{2em}
    L_{surface} = \frac{1}{N} \sum_{i=1}^{N} 1 - sim_i
\end{equation}

Due to noisy and incomplete ground truth plane annotations, we also make use of ground truth depth data, $D \in \mathbb{R}^{H\times W}$, for additional supervision. We calculate the plane induced depth at pixel location $i$ by
\begin{equation}
    D_i^{\, *} = \frac{d_i^{\, *}}{\mathbf{n}_i^{*T} \cdot K^{-1}\mathbf{q}_i} \, ,
\end{equation}
where $K$ represents the ground truth camera intrinsics of the scene and $q_i$ is the x and y index for pixel location $i$. The plane induced depth loss, $L_{depth}$, is formulated as: 
\begin{equation}
    L_{depth} = \frac{1}{N} \sum_{i=1}^{N} | D_i - D_i^{\, *}|.
\end{equation}

We use the plane structure induced loss, first introduced by \citep{planes18} and which we denote by $L_{geom}$, based on the principle that the dot product of a 3D point on a plane with the normal equals the offset, $n^TQ = d$. We use ground truth depth and camera intrinsics to retrieve the 3D point at each pixel location. $Q_i = D_i \, K^{-1}\mathbf{q}_i$ obtains the 3D point projected at one location. 
\begin{equation}
    L_{geom} = \frac{1}{N} \sum_{i=1}^{N} \mathbf{n}_i^{*T} \cdot Q_i - d_i^{\, *}
\end{equation}

\textbf{Gradient Weighting} We add gradient edge weighting as a model variation, weighting $L_{depth}$ and $L_{geom}$ to emphasize learning at edges, areas which are typically more difficult to learn. We choose to use the gradient of the image, $G \in \mathbb{R}^{H\times W}$ rather than depth, in order to better capture edges. Despite more noise at non-edge areas, it can capture more plane edges as some plane instances can have the same depth but still represent different surfaces (\eg picture frame on a wall). This addition results in cross-task improvements for segmentation mask prediction in the case of the multi-view model (see Section \ref{sec:mvfeatwarp}). 
\begin{equation}
    L_{depth, geom} = \frac{1}{N} \sum_{i=1}^{N} G_i * L_i
\end{equation}
The total loss for plane guidance is
\begin{equation}
    L_{P} = L_{plane} + L_{surface} + L_{geom} + L_{depth},
\end{equation}
and the final combined losses:
\begin{equation}
    L_{total} = L_M * w_M + L_C + L_P.
\end{equation}

\subsection{Multiview Plane Feature Guidance}
\label{sec:mvfeatwarp}

\begin{figure}[h]
\centering
\includegraphics[width=0.93\linewidth]{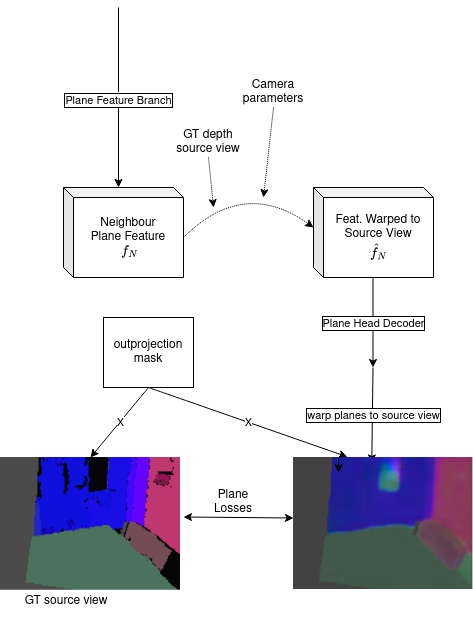} 
\caption{Overview of the feature warping guidance.}
\label{fig:featwarp}
\end{figure}

In this section, we introduce our multi-view guidance approach, depicted in Fig \ref{fig:featwarp}. We take neighbouring image pairs, which we denote by source and neighbouring view $(I_s, I_n)$, and extract the corresponding 2D features. The two finest pyramid feature maps are fused to generate plane features $f \in \mathbb{R}^{\frac{1}{4}H \times \frac{1}{4}W \times C}$. We backproject the neighbouring feature $f_N$ to the corresponding location of the source view using bilinear interpolation. This process uses the ground truth depth, intrinsic parameters, and the relative transform between the views to obtain the warped 2D coordinates, from which we obtain the out-projection mask. We then decode the warped neighbouring feature $\hat{f}_N$ with the plane prediction head to get the corresponding plane parameters. It is important to note that $\hat{f}_N$ contains plane information of the neighbouring view, under the camera coordinates of $I_n$. Therefore, we transform the decoded plane parameters to the source view's camera coordinates before comparing to ground truth. This transformation is given by: 
\begin{equation}
    \mathbf{\hat{n}_s }= \mathrm{R}\, \mathbf{n_n}, \hspace{2em} \hat{d}_s = d_n + \mathbf{n_n}^T \cdot \mathbf{t },
\end{equation}
where $(R, \mathbf{t})$ represents the rotation matrix and translation vector from neighbour to source view, and $(\mathbf{n_n}, d_n)$ are the normal and offset in the neighbouring view. We then calculate an additional plane loss $L_P$ using the transformed plane parameters decoded from the warped feature, excluding from the loss areas that are occluded or fall outside of the 2D image coordinates using the out-projection mask. 

\subsection{Instance Plane Soft-Pooling} To obtain the final instance level plane parameters, we use a soft-pooling technique which only considers per-pixel parameters within the area of the predicted instance. We found that restricting the pooling to this binary area yields better results compared to using soft-pooling across all pixel locations. We opted to not use an instance level plane loss as it negatively impacts the learning of mask segmentation. We generate a binary segmentation mask by applying a threshold to the predicted soft mask, denoted as $m^* \in [0,1]$. The instance level parameter can be retrieved by
\begin{equation}
    \mathbf{p}_{ins} =  \frac{\sum_{i=1}^{M} \, m_i^* \, \mathbf{p}_i^* }{\sum_{i=1}^{M} \, m_i^*},
\end{equation}
where $M$ represents all the pixels falling within the region indicated by the binary segmentation mask, and $\mathbf{p}_i^*$ the predicted plane parameter at the corresponding location.

\section{\uppercase{Experiments}}
We evaluate various configurations of our model as well as comparison models. The nomenclature for our model versions is as follows: SOLOP-5lvls is a single view version using the original 5 feature levels for prediction, SOLOP-SV refers to the single-view model trained on 60,000 samples, SOLOP-MV is the multi-view model trained on 30,000 pairs, and SOLOP-MV-gw incorporates gradient edge loss weighting into the multi-view model. Qualitative results are obtained using the last configuration, as it achieved the best performance.

\subsection{Setup \& Training details}

For comparison between different model versions, we train a base model initialized with a pretrained ResNet-50 backbone and employ a data augmentation scheme where each sample has a 15\% chance of undergoing one of several augmentations, such as a) jitter of brightness, contrast, hue, saturation, b) Planckian jitter \citep{planckian}, c) Gaussian noise, or d) motion blur. We use learning rate warm-up for the first 2000 steps starting from a learning rate of $1\mathrm{e}{-6}$ and increases until $2\mathrm{e}{-4}$. After the initial warm-up period, the learning rate is reduced by a factor of 0.1 given no improvement to the validation loss. For quicker and more fair comparison of model variations, a base model with the best validation loss was saved at epoch 9 and used as initialization to our main models, which were trained for 11 additional epochs. We employ early stopping if validation loss fails to improve for 5 consecutive epochs and save the model with best validation performance as well as the last checkpoint. For evaluation, we take the best of either saved model. The additional models trained using the base model initialization do not use data augmentation, and have 500 steps of learning rate warm-up starting from $1\mathrm{e}{-6}$ to $1\mathrm{e}{-5}$. We use a batch size of 32 for the single view model with gradient accumulation to mitigate the higher instability associated with multi-task learning. We train the models on a single NVIDIA Ampere A100 GPU. For evaluation and FPS calculation, we use a single NVIDIA GeForce RTX 3090 GPU for all models. 

\begin{table*}[!ht]
  \centering
    \caption{Model comparison results on ScanNet dataset for variations of SOLOP model and other single-image planar reconstruction methods.}
     \fontsize{8.8pt}{11pt}\selectfont
     \setlength{\tabcolsep}{3pt}
  \begin{tabular}{@{\extracolsep{0.2pt}} l ccccccc c cc}
    \toprule
    \textbf{Method} & \multicolumn{7}{c}{\textbf{Depth Metrics}}  & \multicolumn{2}{c}{\textbf{Detection Metrics}} & \textbf{FPS} \\
     & AbsRel$\downarrow$ & SqRel$\downarrow$ & RMSE$\downarrow$ & log\_RMSE$\downarrow$ & $\delta < 1.25 \uparrow$ & $\delta^2 < 1.25 \uparrow$ & $\delta^3 < 1.25 \uparrow$ & AP & mAP \\
    \midrule
    PlaneAE & 0.181 & 0.092 & 0.325 & 0.208 & 0.746 & 0.931 & 0.983 & - & - & 17\\
    PlaneTR &   0.178 & 0.133 & 0.365 & 0.215 & 0.768 & 0.930 & 0.977 & - & - & 15 \\
    PlaneRCNN & 0.165 & 0.070 & 0.278 & 0.187 & 0.780 & 0.954 & 0.991 & 0.193 & - & 7 \\
    SOLOP-5lvls$^*$ & 0.143 & 0.059 & 0.276 & 0.185 & 0.813 & 0.960 & 0.990 & 0.416 & 0.314 & 38\\
    SOLOP-SV$^*$ & 0.134 & \textbf{0.052} & \textbf{0.259} & 0.178 & 0.832 & \textbf{0.964} & 0.991 & 0.389 & 0.267 & \textbf{43}\\
    SOLOP-MV$^*$ & 0.136 & 0.054 & 0.261 & \textbf{0.177} & 0.832 & 0.962 & 0.991 &  0.427 & 0.344 & \textbf{43} \\
    SOLOP-MV-gw$^*$ & \textbf{0.133} & \textbf{0.052} & \textbf{0.259} & \textbf{0.177} & \textbf{0.833} & \textbf{0.964} & \textbf{0.992} & \textbf{0.434 }& \textbf{0.347} & \textbf{43} \\
    \bottomrule
     $^*$ = Ours
  \end{tabular}
 
  \label{tab:eval}
\end{table*}

\subsection{Dataset}
For training and evaluation, we use the ScanNet dataset which contains RGB-D images from video sequences, totalling 2.5 million views complete with camera parameters and instance level semantics \citep{scannet}. The ground truth plane instance annotations for instance masks and plane parameters are generated by the authors of PlaneRCNN, and we follow the same process for filtering and preprocessing the planes \citep{planercnn}. We also obtain the corresponding plane instance semantics from the metadata of the plane annotations. 
The ground truth plane data often exhibited issues such as over-segmented, rough edges, or missing plane instances, as planes with a depth error above a $0.1$ meter threshold were omitted. For multi-view guidance training, we take sample pairs which are 10 time-steps away. In some cases, a neighbouring ground truth plane image might contain a segment which is missing in the source view, and vice versa. For the single-view model, we use 60,000 random samples from the training set and 10,000 from the validation set. For the multi-view model, we use 30,000 neighboring pairs for training and 5,000 pairs for validation.
\begin{figure}[!h]
    \centering
    \includegraphics[width=1.07\linewidth]{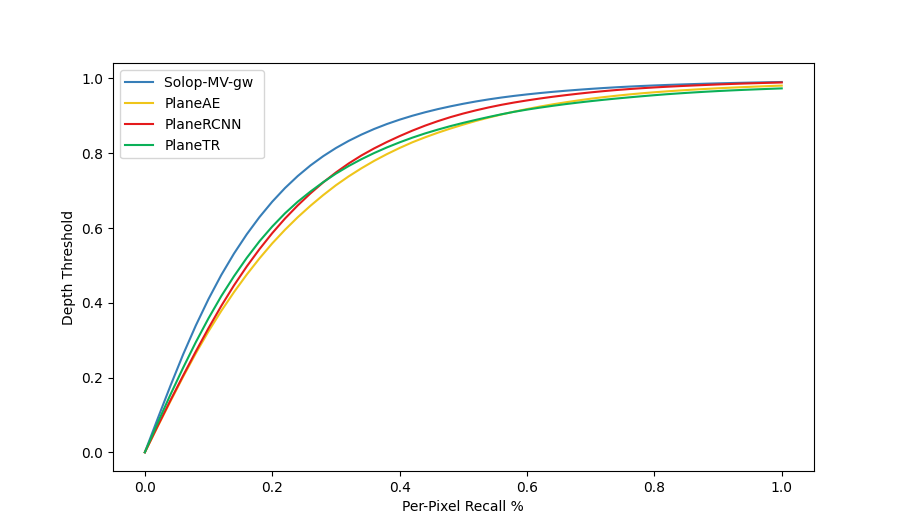}
    \caption{Per-pixel recall at varying depth thresholds in meters.}
    \label{fig:deptheval}
\end{figure}

\subsection{Comparison}
Our model is most comparable to the PlaneAE model from \cite{planeAE} and PlaneTR model from \cite{planetr}, primarily due to the speed of prediction and the fact that they predict plane parameters directly using a single image as input. 
Given the inconsistent quality of ground truth plane data, the authors of PlaneMVS manually selected stereo pairs for the test set, which contained samples with more complete plane annotations \citep{planemvs}. We run our evaluations using the same test set. 
For a fair comparison, we train the PlaneAE model for a total of 20 epochs using a ResNet-50 backbone and the same data with an input size of 480 x 640. The original model was trained using an input size of 192 x 256, resulting in a higher FPS. To align with our training regimen, we train PlaneAE for 11 epochs using 60,000 samples and an additional 9 epochs with 100,000 random samples. We retain the original training configuration of the authors \citep{planeAE}. We use the same approach for retraining the PlaneTR model, and generate the required line segments using HAWPv3 \citep{hawpv3}. While PlaneRCNN also takes a single image at inference time, its slower inference speed makes it a less direct comparison. We run evaluations on the provided model from authors \cite{planercnn}.

\subsection{Evaluation Metrics}
We follow previous methods \citep{planeAE, planercnn} and calculate the per-pixel depth recall at varying thresholds in meters, shown in Fig.~\ref{fig:deptheval}. We also calculate standard depth and detection metrics for a comprehensive evaluation of model performance. Average Precision (AP) is used to assess the quality of the predicted masks, and Mean Average Precision (mAP) takes into account the semantic labels by averaging AP across class categories. For depth metrics, we use Absolute Relative Difference (AbsRel), Squared Relative Difference (SqRel), Root Mean Squared Error (RMSE), log RMSE, and delta accuracy \citep{depthmetrics}. We also calculate model efficiency using Frames Per Second (FPS). The results of these evaluations are summarized in Table \ref{tab:eval}, which shows a marked improvement using our architecture. 

\begin{figure}[ht]
    \centering
        \includegraphics[width=0.493\linewidth]{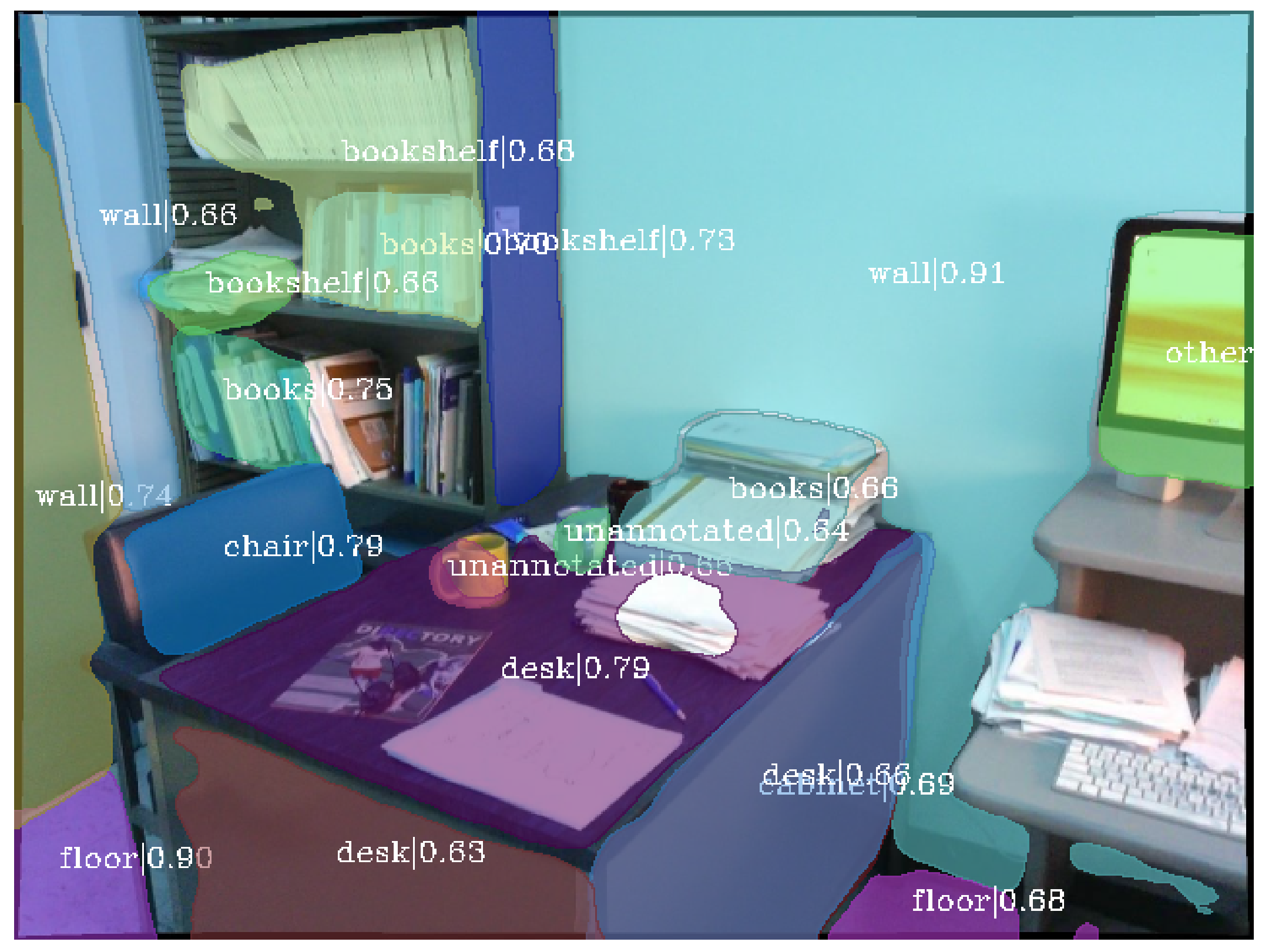}
        \includegraphics[width=0.493\linewidth]{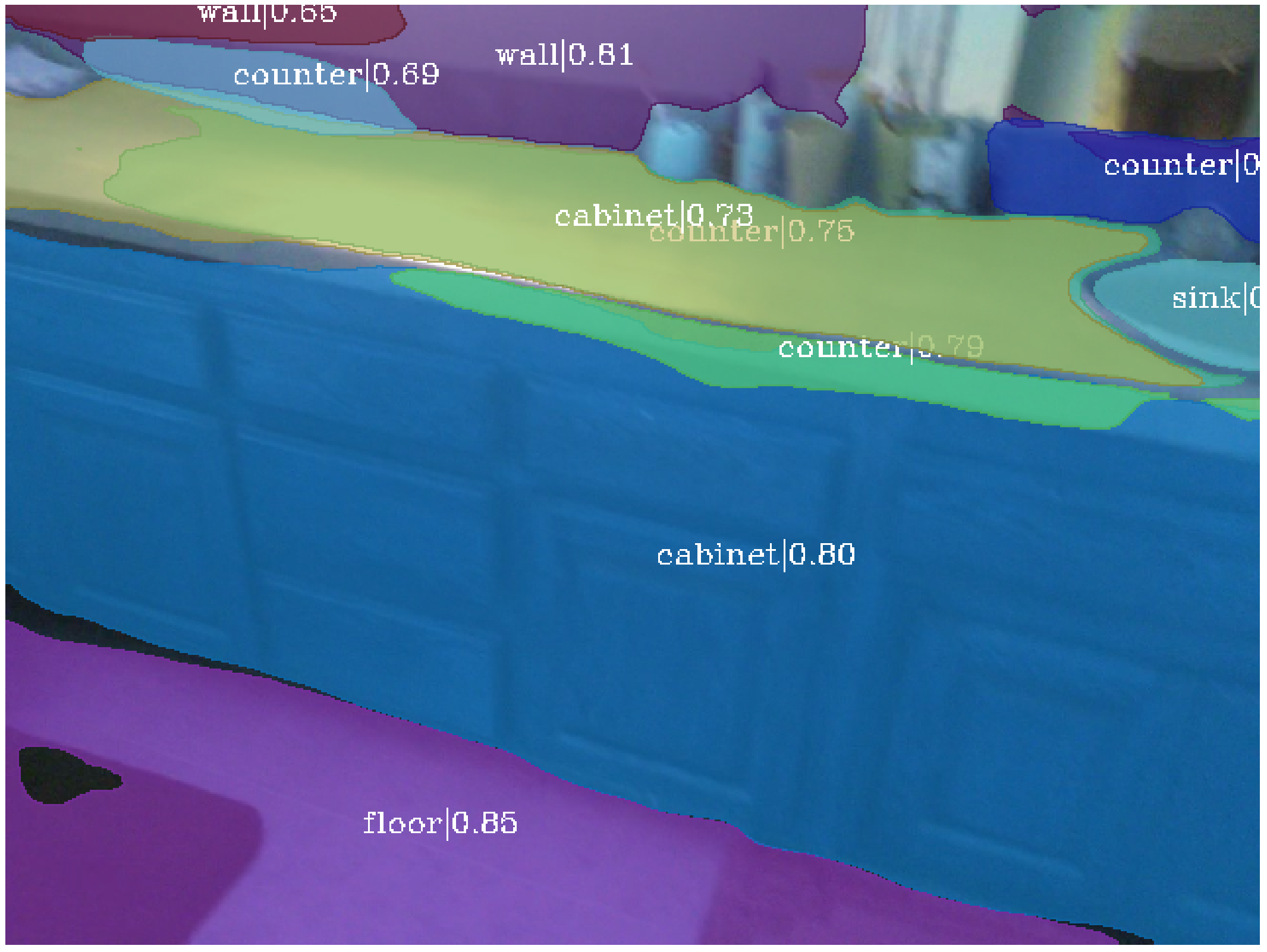}
        \includegraphics[width=0.493\linewidth]{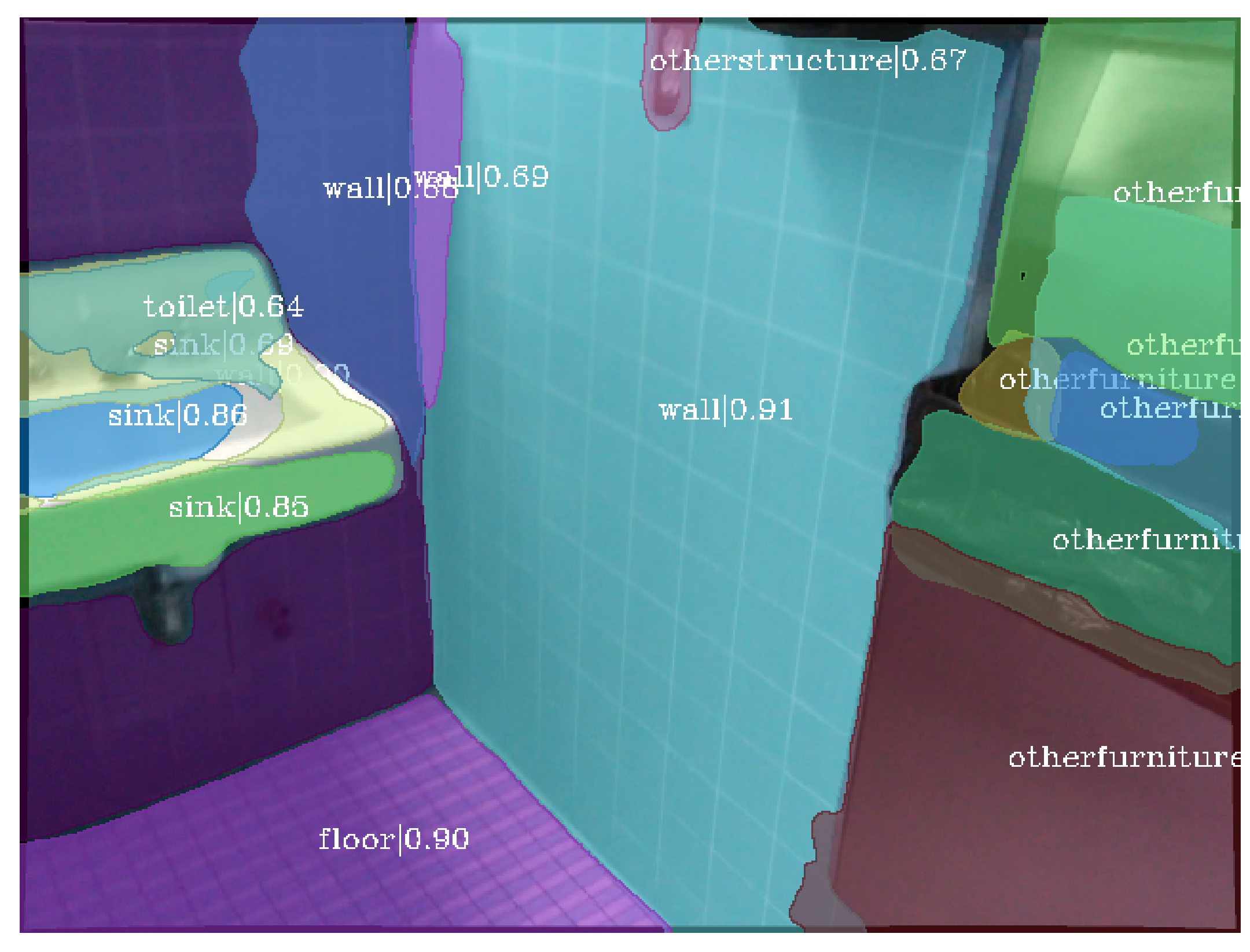}
        \includegraphics[width=0.493\linewidth]{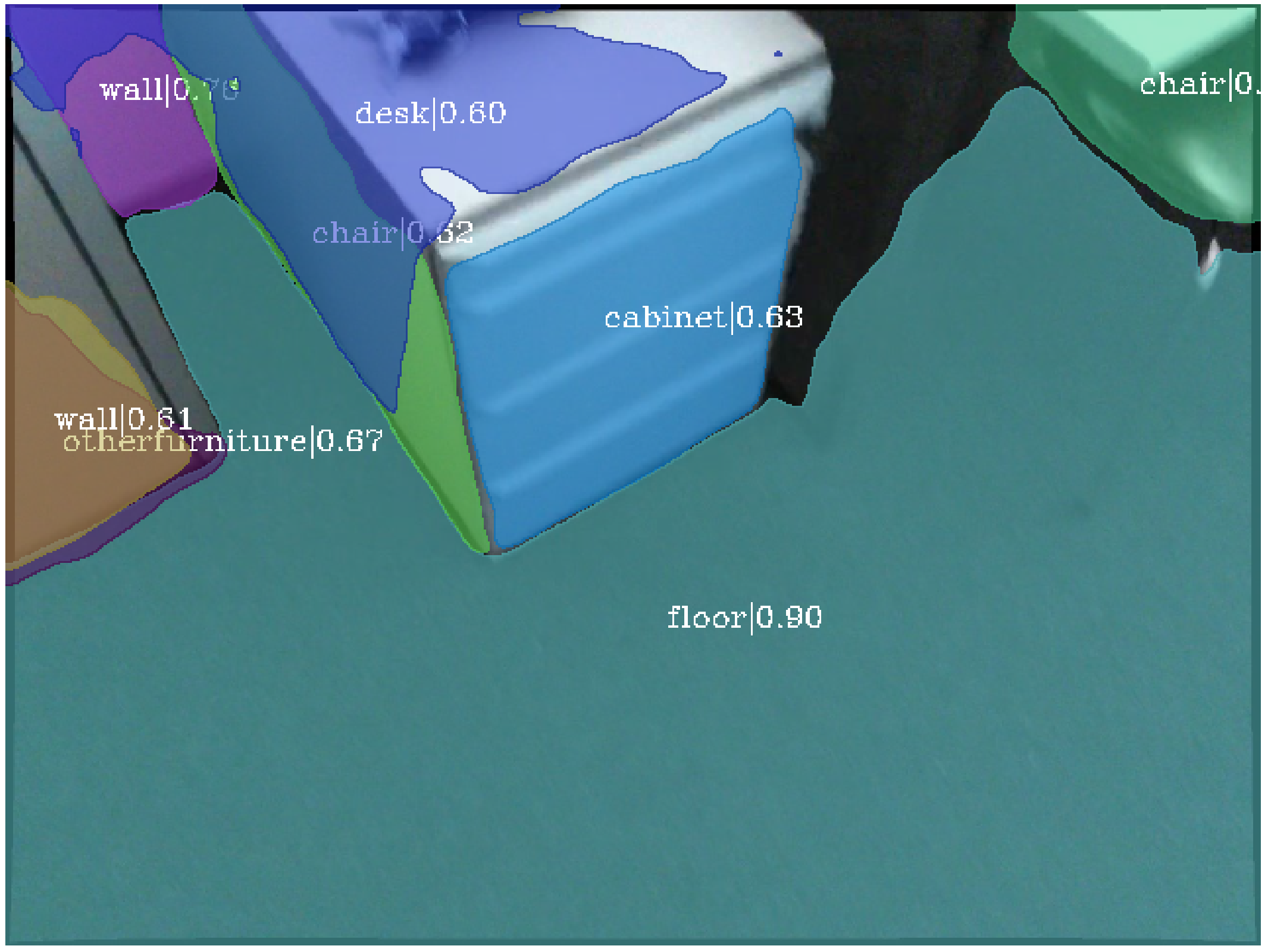}
    \caption{Visualization of semantic predictions using the \mbox{SOLOP-MV-gw} model.}
    \label{fig:sem}
\end{figure}

\subsection{Results}

The task of segmentation becomes more challenging when predicting multiple classes, as overlapping masks from different classes are less likely to be suppressed. The oversegmentation issue appears to be more pronounced in the single view model, whereas multi-view guidance using plane features helped to produce more complete and less over-segmented masks. This improvement is likely attributable to feature sharing and the correlation between ground truth plane instance masks and plane parameters. 

Despite using multi-view guidance on plane predictions, we observe an objective improvement in prediction of segmentation masks. We hypothesize that this is especially effective when adjacent views have disparate ground truth data, such as in the case of missing annotations. This would explain the similar performance with regards to depth metrics between SOLOP-SV and multiview variants, as the ground truth depth is fairly stable across views. Ground truth mask completeness can differ across neighbouring views due to lower quality segments being filtered out. Even though the variants using multi-view guidance saw a lower diversity of scenes compared to the single view version, it nevertheless outperforms the single view variant on the task of mask segmentation. 

\begin{figure*}[!ht]
    \centering
    \includegraphics[width=\textwidth]{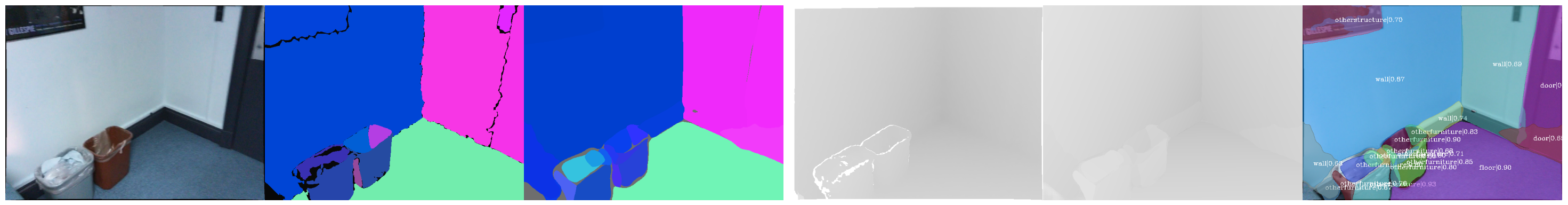}
     \includegraphics[width=\textwidth]{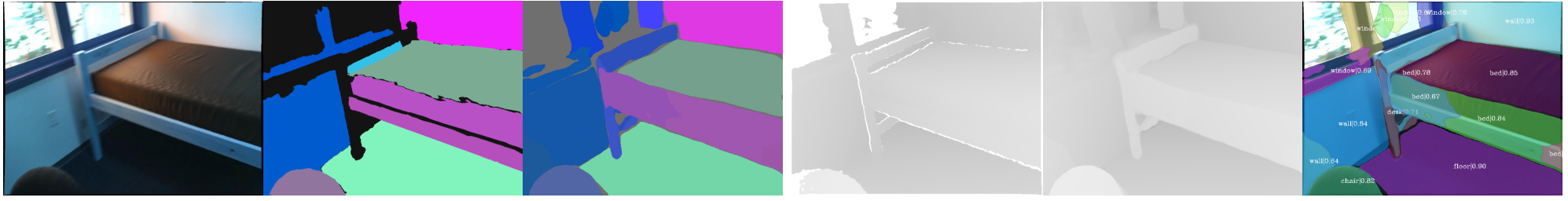}
     \includegraphics[width=\textwidth]{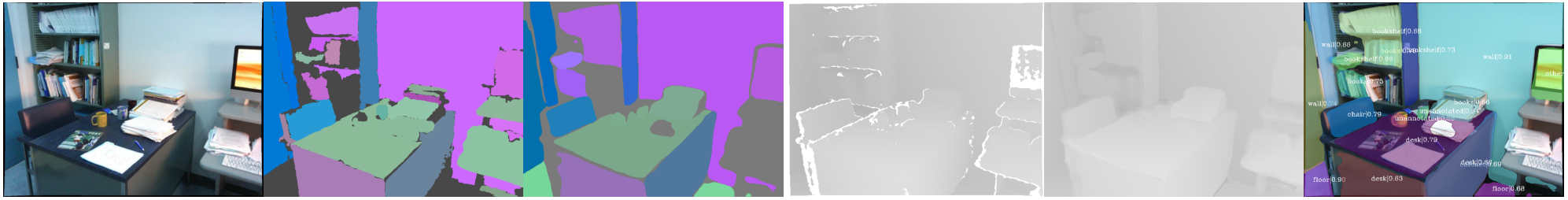}
     \includegraphics[width=\textwidth]{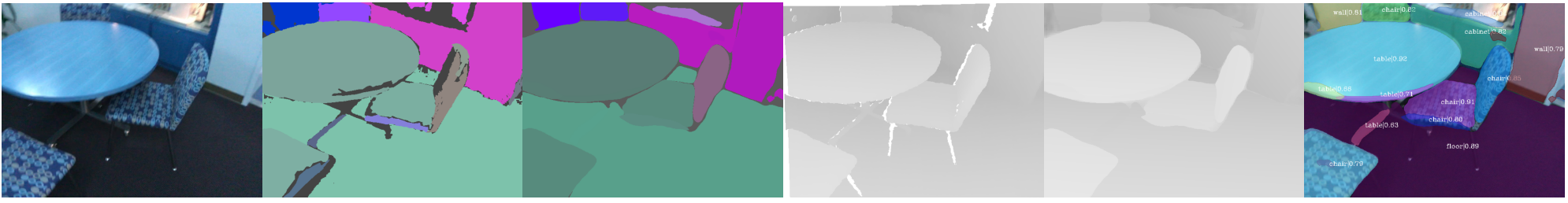}
    \caption{Qualitative results of instance plane and semantic prediction using model with best performance, SOLOP-MV-gw. From left to right: Input image, GT planes, predicted planes, GT depth, predicted depth, predicted semantics.}
    \label{fig:examples}
\end{figure*}

\textbf{Quantitative} All results are obtained using the selected test set chosen by the authors of PlaneMVS. The authors \cite{planemvs} manually selected a higher quality set to evaluate on due to the incomplete and imprecise nature of the ground truth plane annotations. The resulting test set contains 949 image pairs. 
Our quantitative findings from model comparisons, summarized in Table \ref{tab:eval}, indicate that our multi-view model variant not only matches the performance of the single-view model in depth metrics, but also shows a significant improvement in detection metrics. This demonstrates the efficacy and improved data efficiency in using multi-view guidance via warping in feature space, at least in the case of using shared features for multitask learning. Since all SOLOP variants use a single image at inference time, the FPS result is the same for the versions of the model using 3 feature levels (SOLOP-SV, SOLOP-MV, SOLOP-MV-gw), but significantly reduced for the version with the original 5 level architecture (SOLOP-5lvls).
SOLOP-MV-gw achieves better depth recall comparatively (see Fig.~\ref{fig:deptheval}), while all SOLOP variants outperform the comparison models on standard metrics.  


\textbf{Qualitative} We display different types of visual results from our best model in Figures \ref{fig:teaser}, \ref{fig:sem}, and \ref{fig:examples}. In contrast to previous works that predicted a binary plane indication, the incorporation of multi-class semantics introduces an added complexity. The change made to the focal loss for category predictions (see Section \ref{sec:losses}) leads to more confident scoring as well as a potential increase in false positives, which is already exacerbated in the case of multi-class predictions. However, we found that raising the score threshold for the final masks partially mitigated this issue. 
See Fig.~\ref{fig:examples} for visual results. The structure of the scene is easier to predict than the exact depth, a challenge presented when using a single image for inference. 
Sample visualizations of the semantic predictions can be found in Fig.~\ref{fig:sem}. Cases of oversegmentation can occur due to prediction of different classes, or different plane orientation, as each mask represents a planar segment associated with a class label. Overall, our model demonstrates robust performance both visually and quantitatively for the task of planar reconstruction with semantic labels. 

\section{\uppercase{Discussion and Future Work}}
In this work, we introduce SOLOPlanes, a real-time semantic planar reconstruction network which shows cross-task improvement when using multi-view guidance in feature space. The task of predicting plane parameters from a single image is non-trivial, and the complexity is further compounded by multi-task learning. Despite these challenges, our model competes favorably with other, less efficient methods in planar reconstruction that do not offer semantic predictions. To the best of our knowledge, our model also outperforms all other planar reconstruction models in computational efficiency, measured using FPS. Our work advances semantic plane instance segmentation without sacrificing computational efficiency, striking a balance between efficiency and performance. We hope it will serve as an inspiration or stepping stone for further research geared towards applications with real-world impact.  

\clearpage

{\small
\bibliography{example}}



\end{document}